\documentclass[letterpaper, 10 pt, conference]{format/ieeeconf}  

\IEEEoverridecommandlockouts                              

\usepackage{dsfont}
\usepackage{amsmath}

\usepackage{graphics} 
\usepackage{amsmath} 
\usepackage{amssymb}  
\usepackage{todonotes}

\usepackage{amsfonts}
\usepackage{mathtools}
\usepackage[linesnumbered,algoruled,boxed,vlined, noend]{algorithm2e}
\usepackage{amsthm}
\usepackage{comment}
\usepackage{cite}
\usepackage{wrapfig}
\let\labelindent\relax
\usepackage{enumitem}
\usepackage{overpic}

\DeclareMathAlphabet{\mathcal}{OMS}{cmsy}{m}{n} 

\newtheorem{proposition}{Proposition}

\theoremstyle{definition}

\theoremstyle{remark}

\DeclarePairedDelimiterX{\norm}[1]{\lVert}{\rVert}{#1}

\newcommand{\rom}[1]{\uppercase\expandafter{\romannumeral #1\relax}}

\newcommand{\disk}{\text{$\mathcal{D}$}}

\newcommand{\arrangement}{\text{$\mathcal{A}$}}
\newcommand{\workspace}{\text{$\mathcal{W}$}}
\newcommand{\objects}{\text{$\mathcal{O}$}}

\newcommand{\DFSDP}{DFS_{DP}\xspace}
\newcommand{\EDFSDP}{EDFS_{DP}\xspace}

{\end{list}} 
\usepackage{soul} 
\usepackage[a-2b,mathxmp]{pdfx}[2018/12/22]

\title{\LARGE \bf
Uniform Object Rearrangement: From Complete Monotone Primitives to Efficient Non-Monotone Informed Search}

\author{Rui Wang$^{*}$, Kai Gao$^{*}$, Daniel Nakhimovich$^{*}$, Jingjin Yu and Kostas E. Bekris%
\thanks{$^{*}$The first three authors contributed equally to this paper. The authors are with the Dept. of Computer Science, Rutgers Univ., NJ. Email: {\tt\small { \{rw485, kg627, dn332, jy512, kb572\}}@cs.rutgers.edu}.
The work is supported in part by NSF awards IIS-1845888, NRI-1734492, CCF-1934924 and an NSF NRT project 2021628.
}%
}

\begin{document}

\maketitle
\thispagestyle{empty}
\pagestyle{empty}


\begin{abstract} 
Object rearrangement is a widely-applicable and challenging task for robots. Geometric constraints must be carefully examined to avoid collisions and combinatorial issues arise as the number of objects increases. This work studies the algorithmic structure of rearranging uniform objects, where robot-object collisions do not occur but object-object collisions have to be avoided. The objective is minimizing the number of object transfers under the assumption that the robot can manipulate one object at a time. An efficiently computable decomposition of the configuration space is used to create a ``region graph'', which classifies all continuous paths of equivalent collision possibilities. Based on this compact but rich representation, a complete dynamic programming primitive $\tt \DFSDP$ performs a recursive depth first search to solve monotone problems quickly, i.e., those instances that do not require objects to be moved first to an intermediate buffer. $\tt \DFSDP$ is extended to solve single-buffer, non-monotone instances, given a choice of an object and a buffer. This work utilizes these primitives as local planners in an informed search framework for more general, non-monotone instances. The search utilizes partial solutions from the primitives to identify the most promising choice of objects and buffers. Experiments demonstrate that the proposed solution returns near-optimal paths with higher success rate, even for challenging non-monotone instances, than other leading alternatives.

\end{abstract}

\section{Introduction}

\begin{figure}[t]
    \centering
    \includegraphics[width=0.19\textwidth]{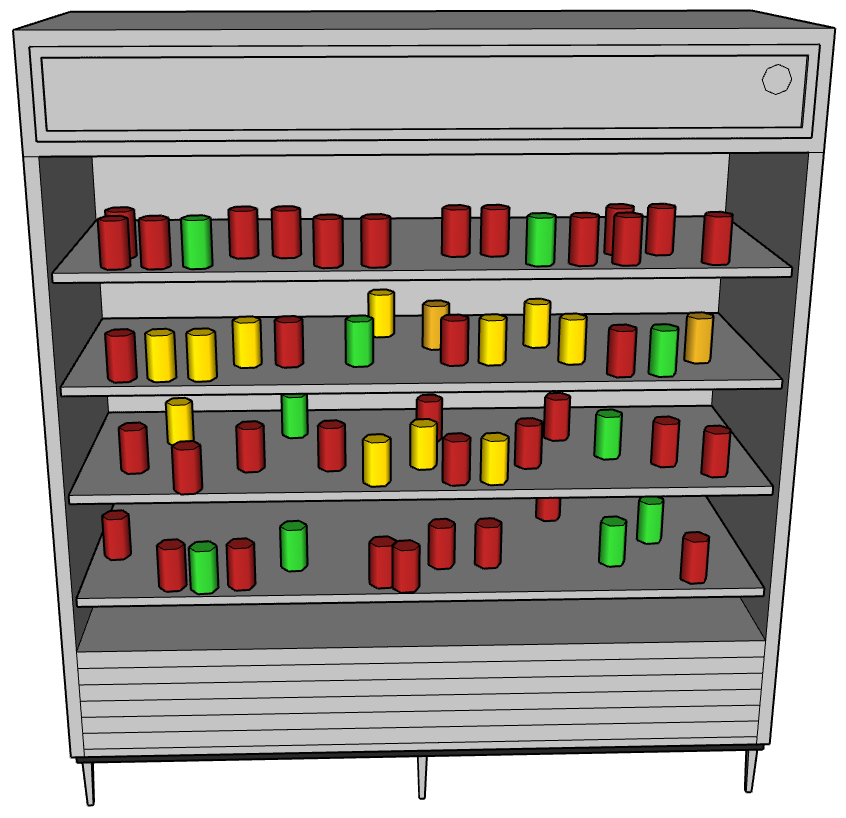}
    \hspace{0.1in}
    \includegraphics[width=0.23\textwidth]{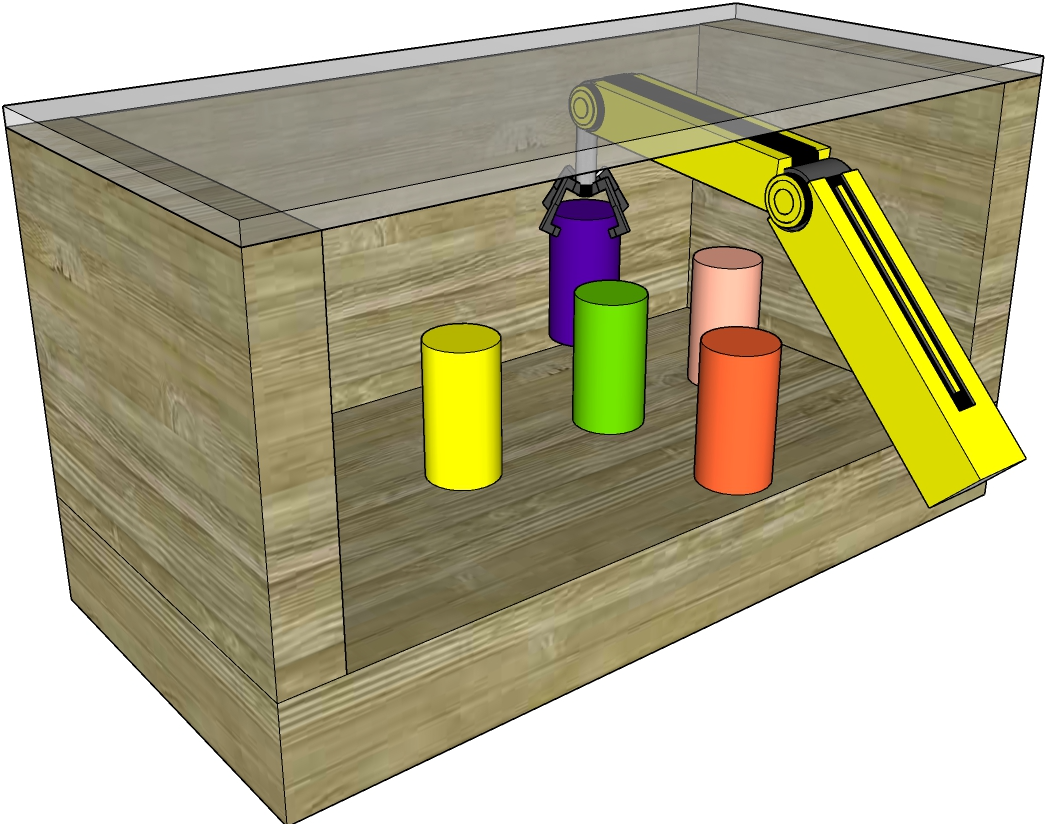}
    \vspace{-.05in}
    \caption{(left) Robots in logistics have to rearrange similar objects in shelves. (right) The focus is on combinatorial and geometric aspects of rearrangement when the arm can reach objects without colliding with them but cannot lift objects to guarantee they do not collide with each other.}
    \label{fig:shelf_example}
\end{figure}

Object rearrangement is a critical robot skill broadly applicable in the logistics, industrial, and service domains. For instance, robots can rearrange merchandise in grocery shelves as in Fig.~\ref{fig:shelf_example}(left), retrieve food in packed fridges for home automation, or perform packaging of products for shipping \cite{han2019toward}.  This work focuses on problems where one object is manipulated at a time without incurring any object-object collisions. The objective is to minimize the number of object transfers needed to complete a rearrangement. The setting is akin to an attendant moving cars in a crammed parking lot. Clearly, collisions between two cars should not occur and the attendant should minimize the number of times he drives a car. Similar scenarios occur in manipulation, e.g., when a soda must be retrieved from a fridge, multiple other beverages may have to be rearranged first. While the arm may be able to reach the objects with an overhand grasp as in Fig. \ref{fig:shelf_example}(right), it may not be possible to just lift objects to avoid collisions among them. Furthermore, given the sometimes unpredictable effects of object-object collisions, these collisions should be avoided. 


\begin{figure*}[ht]
    \centering
    \includegraphics[width=0.95\textwidth]{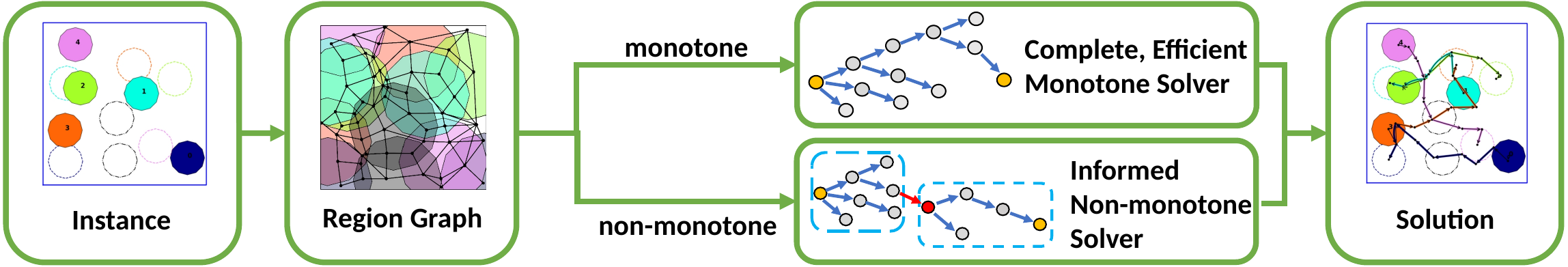}
    \vspace{-.1in}
    \caption{A rearrangement instance is defined by the objects' starts and goals. The approach pre-allocates candidate buffers and decomposes the space to generate a region graph. Each region corresponds to a different interference set with existing start, goal and buffer locations. If a monotone solution exists given all object paths in the region graph, a dynamic programming approach discovers it efficiently. For non-monotone problems, a search method incrementally builds a tree, where nodes are arrangements and edges represent object transfers to their goals together an object moved to a buffer (called a ``perturbation''). Informed heuristics based on the dynamic programming solver guide the choice of objects and buffers for faster, high-quality solutions.}
    \label{fig:pipeline diagram}
\end{figure*}

A rearrangement domain, where performance guarantees can be argued but which is already hard, involves tabletop setups and overhand grasps, where both robot-object and object-object collisions can be ignored \cite{han2018complexity}. This work pushes forward the understanding of the algorithmic structure of rearrangement by considering object-object collisions for uniform-shaped objects in planar setups. A principled solution pipeline is proposed (Fig.~\ref{fig:pipeline diagram}), which yields complete and efficient primitives for monotone instances and an effective informed search framework that quickly computes high-quality non-monotone solutions. In a monotone instance, each object needs to be moved at most once, i.e., without having to move first to an intermediate, buffer location. The pipeline is composed of three key components/contributions:

\noindent 1. A decomposition of the space into equivalent regions resulting in a compact \emph{region graph} containing all possible object paths in terms of collision sets that can arise. It abstracts away the problem's continuous, geometric aspects.

\noindent 2. A novel \emph{dynamic programming} routine $\tt \DFSDP$ that solves monotone instances quickly by reasoning over object paths on the region graph. It's extended to optimally solve 1-buffer, non-monotone problems for an object and buffer choice. 

\noindent 3. An \emph{informed search} framework that  uses $\tt \DFSDP$ variations as local planners. Given partial solutions generated by $\tt \DFSDP$, it generates heuristics to explore promising objects and buffers for quickly making progress towards the goal arrangement. 

Simulation experiments show the efficiency of the monotone solver $\tt \DFSDP$. The informed search is shown to be faster and to return higher-quality solutions than alternatives for non-monotone instances. For problems with 1 or 2 buffers where the optimal solution can be discovered, the framework returns almost optimal solutions. An ablation study compares against a baseline search that uses the monotone solver as a local planner. It highlights the benefits of the non-monotone extension of $\tt \DFSDP$ and of the chosen heuristics.

\label{sec:intro}

\section{Related Work}

An apparently simple setup involves picking objects from tabletops, which are lifted sufficiently high before being placed back to avoid object-object collisions \cite{han2018complexity, shome2018fast, havur2014geometric, dabbour2019placement}. This setting is as computationally hard to solve optimally as the Travelling Salesperson Problem (TSP) when objects' starts and goals do not overlap \cite{han2018complexity}. Overlapping starts and goals complicates the problem as it reduces to the Feedback Vertex Set (FVS) problem \cite{Kar72}, which is possibly APX-hard. Dual arm rearrangement allows for parallelism but complicates reasoning \cite{shome2018fast}. Integer programming is often applied for deciding the object order together with motion planning \cite{han2018complexity, shome2018fast}. Optimal tabletop placement has also been approached via Answer Set Programming (ASP) \cite{havur2014geometric, dabbour2019placement} and informed heuristics \cite{cosgun2011push}. Similarly to this work, these efforts aim to minimize object grasps/pushes but address packing new items into cluttered environments; this is more akin to the unlabeled version of this works' problem. 


For more confined spaces, small monotone problems have been addressed via backtracking search \cite{stilman2007manipulation}. A useful structure for solving problems in the general case is a ``dependency graph'' \cite{van2009centralized}, which expresses constraints between objects given their starts and goals. Acyclic graphs indicate existence of monotone solutions. The ``true'' dependency graph is difficult to construct in general as all object paths must be considered. With approximations for Minimum Constraint Removal (MCR) paths \cite{hauser2014minimum}, it is possible to practically build good dependency graphs \cite{krontiris2015dealing, krontiris2016efficiently} to solve general instances. This paper focuses on uniform disc-shaped object instances and object-object interactions. By constructing a ``region graph'' that compactly represents all object paths, it is possible to search over dependency graphs with efficient and complete solvers for monotone and non-monotone instances.

Solutions for integrated task and motion planning (TAMP) can be applied to general rearrangement \cite{garrett2015ffrob, srivastava2014combined}. While they incorporate heuristics and are probabilistically complete (PC), it is difficult to make arguments about optimality. Furthermore, insights from rearrangement planners can lead to effective heuristics for TAMP. Pushing allows simultaneous action on multiple objects \cite{ben1995push,ben1998practical,huang2019large} though actions may be irreversible. Pushing has been studied in the context of robust rearrangement under uncertainty \cite{dogar2012planning, koval2015robust, anders2018reliably}.


Navigation Among Movable Obstacles (NAMO) \cite{chen1990practical} is related to rearrangement. It is NP-hard  \cite{stilman2008planning} and its difficulty depends on linearity and monotonicity notions. A problem is linear if collision free components can be traversed in sequence, where earlier actions do not constrain future ones. For non-monotone, non-linear NAMO problems, a PC algorithm exists for axis-aligned objects and robots \cite{van2009path} though it may return highly-redundant paths. Recently, methods have tackled online NAMO settings \cite{wu2010navigation, kakiuchi2010working, levihn2014locally}. In object retrieval, movable obstacles may obstruct paths and works have used dependency reasoning to generate valid plans \cite{dogar2012planning,dogar2014object} and scale linearly in actions with the number of objects \cite{danielczuk2019mechanical}. Recently, algorithms have been proposed to explicitly minimize the number of objects to relocate \cite{nam2019planning, nam2020fast}.



\section{Problem Setup and Notation}
\label{section: problem setup}
Let $\workspace \subset \mathbb{R}^2$ be a bounded polygonal region where $n$ labeled uniform-shaped objects $\objects=\{o_1, \cdots, o_n\}$ reside. An arrangement $\arrangement$ of $\objects$ is given as $(p_1, \cdots, p_n) \in \workspace^n$, where $p_i \in \workspace$ defines the position of $o_i$, i.e., the coordinates of $o_i$'s center. $\arrangement[o_i]=p_i$ indicates that object $o_i$ assumes  position $p_i$ in the arrangement $\arrangement$. Define as $V(p)$ the 
subset of $\workspace$ occupied by an object at position $p$. An arrangement $\arrangement$ is \emph{feasible} if no object-object collisions occurs, i.e., $\arrangement$ is feasible if $\forall i,j \in [1,n], i \neq j: V(\arrangement[o_i]) \cap V(\arrangement[o_j]) = \emptyset$.

A robotic arm can reach objects at any position $p \in \workspace$ without colliding with them. Given an arrangement $\arrangement$, the arm can move one object at a time from its current position $p_i = \arrangement[o_i]$ to a new position $p^{\prime}_i$, giving rise to a new arrangement $\arrangement^{\prime}$, where $\arrangement^{\prime}[o_i] = p^{\prime}_i$ and $\forall j \in [1,n], j \neq i: \arrangement^{\prime}[o_j] = \arrangement[o_j]$. The arm's motion results in continuous paths $\pi_i: [0,1] \to \workspace$ for object $o_i$ with $\pi_i(0) = p_i$ and $\pi_i(1) = p^{\prime}_i$. The arm cannot raise the picked object far enough to guarantee collision avoidance with other objects. Consequently, object paths can be split into valid and non-valid paths. A path $\pi$ for moving object $o_i$ is \emph{valid} if it does not result in a collision between $o_i$ and all the other objects $o_j$ given their positions in a feasible $\arrangement$, where $1 \leq j\leq n, j \neq i$. 


A candidate new position $p \in \workspace$ for object $o_i$ given arrangement $\arrangement$, may cause object $o_i$ to collide with other objects. The set of objects that collide with $o_i$ given its position $p$ is called as an \emph{interference set} for $p$. Specifically, the interference set for position $p$ of object $o_i$ given arrangement $\arrangement$ is defined as $\mathbf{I}_{i,\arrangement}(p) = \{o_j \in \objects, o_j \neq o_i: V(p) \cap V(\arrangement[o_j]) \neq \emptyset \}$. This notion can be extended to a set of arrangements $\mathbf{A}$ as $\mathbf{I}_{i,\mathbf{A}}(p) = \bigcup_{\arrangement \in \mathbf{A}} \mathbf{I}_{i,\arrangement}(p)$.  Furthermore, given a path $\pi_i$ for object $o_i$, the union of interference sets along the path $\pi_i$ contains all objects that $o_i$ will collide with along $\pi_i$. This is defined as the interference set $\mathbf{I}_{i,\arrangement}(\pi_i)$ of path $\pi_i$, i.e., $\mathbf{I}_{i,\arrangement}(\pi_i) = \bigcup_{\forall p \in \pi_i} \mathbf{I}_{i,\arrangement}(p)$.

The rearrangement problem considered here is to discover a sequence of valid object paths which bring objects $\objects$ from an initial feasible arrangement $\arrangement_I$ to a final feasible arrangement $\arrangement_F$. The concatenation of such valid paths gives rise to a solution path sequence $\Pi$. Optimal solution path sequences $\Pi^*$ minimize the number of object paths needed to solve a rearrangement problem. 

This definition transforms the rearrangement problem into a path planning problem on the arrangement space. An \emph{arrangement space} is the space of all feasible object arrangements. For arbitrary arrangements $\arrangement_a$, $\arrangement_b$, there is a transition from $\arrangement_a$ to $\arrangement_b$ if $\exists ! o_i\in \objects \text{ s.t. } \arrangement_a[o_i]\neq\arrangement_b[o_i]$, and there is a valid path $\pi_i$ for $o_i$ from $\arrangement_a[o_i]$ to $\arrangement_b[o_i]$.

An instance is \emph{monotone} if there is a solution path sequence from $\arrangement_I$ to $\arrangement_F$ that contains at most one valid path $\pi_i$ for each $o_i$. \emph{Non-monotone} instances require solution path sequences where at least one object is moved at least twice, i.e., where the object is first placed to an intermediate \emph{buffer} position before moved to its target.

\section{Region Graph}
\label{sec:region_graph}



Given a set of arrangements $\mathbf{A}$ (e.g., $\{\arrangement_I$, $\arrangement_F\}$), the workspace $\workspace$ can be decomposed into a set of closed, path-connected regions $\mathcal{R_\mathbf{A}}$. Each region contains positions with the same interference set. Denote as $\disk(p):=\{p':V[p]\cap V[p']\neq \emptyset \}$ the subset of $\workspace$ for which an object placement will collide with an object located at $p$. Alg. \ref{alg:region_decomp} shows the region decomposition process based on $\mathbf{A}$. For each position $p$ in each arrangement in $\mathbf{A}$, each existing region $r \in R_\mathbf{A}$ is split by $r\cap\disk(p)$ and $r\backslash \disk(p)$ (lines 5-10). After that, regions that are not path-connected are split into their path-connected components (lines 11-13).

 
    


\begin{algorithm}
\caption{Region Decomposition($\mathbf{A}$)}
\label{alg:region_decomp}
\small
\DontPrintSemicolon
$M \gets \emptyset$,
$\mathcal{R_\mathbf{A}} \gets \emptyset$\;
\For{$\arrangement \in \mathbf{A}$}{
\For{$p \in \arrangement$}{
    \For{$r \in \mathcal{R_\mathbf{A}}$}{
        \If{$r \cap \disk(p) \neq \emptyset$}{
            $\mathcal{R_\mathbf{A}} \gets \mathcal{R_\mathbf{A}} \cup \{r \cap \disk(p)\}$,         $r \gets r \setminus \disk(p)$\;
 
            \lIf{$r \text{ is } \emptyset$}{
                $\mathcal{R_\mathbf{A}} \gets \mathcal{R_\mathbf{A}} \setminus \{r\}$
            }
        }
    }
    $d \gets (\workspace \cap \disk(p)) \setminus M$\;
    
    \lIf{$d \neq \emptyset$}{
        $\mathcal{R_\mathbf{A}} \gets \mathcal{R_\mathbf{A}} \cup \{d\}$
    }
    $M \gets M \cup \disk(p)$\;
}
}

\For{$r \in \mathcal{R_\mathbf{A}}$}{
    $\mathcal{R_\mathbf{A}} \gets \mathcal{R_\mathbf{A}} \setminus \{r\}$\;
    \lFor{$c \in \textbf{components}(r)$}{
        $\mathcal{R_\mathbf{A}} \gets \mathcal{R_\mathbf{A}} \cup \{c\}$
    }
}
\KwRet{$\mathcal{R_\mathbf{A}}$}\;
\end{algorithm}




The \emph{region graph} $G_\mathbf{A} = (\mathcal{R_\mathbf{A}}, E_\mathbf{A})$ has regions as nodes. An edge $(r_1, r_2)\in E_\mathbf{A}$ exists if and only if $r_1 \cup r_2$ is a path-connected subset of $\workspace$. Given a walk $W$ on $G_\mathbf{A}$ with vertex sequence $R=(r_0,...,r_k)$, its interference set is $\mathbf{I}(W) = \bigcup_{r_i \in R_\mathbf{A}}\mathbf{I}(r_i)$. Any continuous path in the workspace can be associated with a walk on the region graph with the same interference set. Since multiple paths in $\workspace$ may correspond to the same path in $G_\mathbf{A}$, the paths in the region graph form equivalence classes of workspace paths sharing the same interference set. Although multiple paths in the region graph can have the same interference set, a region graph for one or more arrangements categorizes all paths for an object into a finite number of classes. Thus, for any object, it is possible to systematically explore all path options between two positions in terms of interference sets.


Since the focus is on finding solutions with minimum interference sets, it is not necessary to discover every homotopy class of workspace paths traversing the regions but rather only the classes of paths with varying interference sets.
Thus, in order to reduce the number of region adjacencies, it is reasonable to only check adjacency between regions pairs with interference sets differing by one object. 




\section{High Performance Monotone Solver}
\label{sec: monotone solver}

For a monotone problem, there are $2^n$ different arrangements, where an object can be placed either at its start or goal. Given an arrangement $\arrangement$, let $\objects(\arrangement) = \{o_i:\arrangement[o_i] = \arrangement_F[o_i]\}$ denote the subset of objects at their goals. $\arrangement$ is \emph{accessible} if the subproblem of moving $\objects(\arrangement)$ from $\arrangement_I$ to $\arrangement$ is monotone.
If $\arrangement$ is known to be accessible, solving the subproblem of moving $\objects\backslash\objects(\arrangement)$ from $\arrangement$ to $\arrangement_F$ is sufficient for solving the full problem. In addition, the subproblem's solution does not depend on $\objects(\arrangement)$'s ordering. 

\begin{algorithm}
\DontPrintSemicolon
\begin{small}
    \SetKwInOut{Input}{Input}
    \SetKwInOut{Output}{Output}
    \SetKwComment{Comment}{\% }{}
    \caption{$\tt \DFSDP$($T$, $\arrangement_C$, $\arrangement_F$, $G_{\{\arrangement_I,\arrangement_F\}}$)}
		\label{alg:DFSDP}
    \SetAlgoLined
        \For{$o\in \objects\backslash\objects(\arrangement_C)$}{
            $\arrangement_{new}[\objects\backslash \{o\}]=\arrangement_{C}[\objects\backslash \{o\}]$\\
            $\arrangement_{new}[o]=\arrangement_{F}[o]$\\
            \If{$\arrangement_{new}$ not in $T$}{
                $\pi, D_{\pi} \leftarrow$ RG-DFS($G_{\{\arrangement_I\arrangement_F\}}$, $\arrangement_C$, $\arrangement_{new}$, $D_{\pi}$)
            
                \If{$\pi \neq \emptyset$}{
                    $T[\arrangement_{new}].parent\leftarrow \arrangement_{C}$\\ 
                    \lIf{$\arrangement_{new} \neq \arrangement_F$}{
                        \\
                        T, $D_{\pi}$ = $\tt \DFSDP$(T, $D_{\pi}$, $\arrangement_{new}$, $\arrangement_F$, $G_{\{\arrangement_I,\arrangement_F\}}$)}
                    \lIf{$\arrangement_F \in T$}{\Return T, $D_{\pi}$}
                }
            }
        }
        \Return T, $D_{\pi}$\\
\end{small}
\end{algorithm}

Given this observation, a dynamic program $\tt \DFSDP$ is presented in Alg.~\ref{alg:DFSDP} to solve the monotone problem. It grows a search tree $T$ in the arrangement space rooted at $\arrangement_I$. Each node on the tree is accessible from the root and $\tt \DFSDP$ tests its connection to $\arrangement_F$ in a depth-first manner. Alg.~\ref{alg:DFSDP} first enumerates all possible objects and attempts to move each of them from their current positions to their goals (lines 1-3). When the newly constructed arrangement $\arrangement_{new}$ is not in $T$ (line 4), a valid path for the object $o$ from $\arrangement_C[o]$ to $\arrangement_F[o]$ is searched (line 5). If a valid path is found (line 6), $\arrangement_{new}$ is labeled accessible and added to $T$ (line 7). If $\arrangement_{new}$ is not $\arrangement_F$, the program recurses on it (lines 8-9). Otherwise, the solution is found (line 10) with corresponding object paths.

$\tt \DFSDP$ searches for an object path local to $\arrangement_C$ within a subgraph of the region graph $G_{\{\arrangement_I, \arrangement_F\}}= (\mathcal{R_{\{\arrangement_I, \arrangement_F\}}}, E_{\{\arrangement_I, \arrangement_F\}})$ constructed from $\arrangement_I$ and $\arrangement_F$.  This subgraph ignores dependencies from non-occupied positions. Since there are $2^n$ different region subgraphs (one per arrangement $\arrangement_C$), but only $n$ pairs of start/goal positions, paths are stored in a dictionary $D_{\pi}$ for each start/goal pair for future queries. Since the dictionary $D_{\pi}$ is enriched incrementally, time is saved in subsequent calls to $\tt \DFSDP$ by first checking $D_{\pi}$. RG-DFS, the path finding algorithm for objects on the region graph, is only executed when no valid object path local to $\arrangement_C$ was previously found.

\vspace{-.05in}
\begin{proposition}
$\tt \DFSDP$ over the region graph $G_{\{\arrangement_I, \arrangement_F\}} = (\mathcal{R_{\{\arrangement_I, \arrangement_F\}}}, E_{\{\arrangement_I, \arrangement_F\}})$ is complete for monotone instances.
\end{proposition}
\vspace{-.05in}

Since there exists a one-to-many mapping from region graph paths to all possible object paths in $\workspace$,
the path finding algorithm RG-DFS effectively searches for object paths when searching the region graph in a depth-first manner.
Furthermore, the search tree is finite since the region graph is finite given a finite number of objects.
Therefore, RG-DFS is complete, which leads to the completeness of $\tt \DFSDP$.

In order to solve non-monotone instances, at least one object has to be moved to a buffer. This work refers to an action of moving object $o_i$ to a buffer $b_i$ as a \emph{perturbation} and denoted as $P(o_i, b_i)$. A perturbation at $\arrangement$ splits the movement of object $o_i$ into two phases: $\arrangement[o_i] \rightarrow b_i$ given the perturbation and $b_i \rightarrow \arrangement_F(o_i)$. While a monotone solution can be treated as a permutation of $n$ actions $a_{o_1}, ..., a_{o_n}$ where each action $a_{o_i}$ moves $o_i$ from $\arrangement_I[o_i]$ to $\arrangement_F[o_i]$, a non-monotone solution with the perturbation $P(o_i, b_i)$ can be viewed as a permutation of $n+1$ actions with partial order enforced between the two actions involving $o_i$. Thus, 
the monotone planner $\tt \DFSDP$
is extended
into a planner $\tt \EDFSDP$, which solves 1-buffer, non-monotone problems given a specific choice of object $o_i$ and buffer $b_i$ to which $o_i$ moves.

\section{Non-monotone Framework}
\label{non-monotone framework}


For general non-monotone problems, solution quality and computation time are largely determined by the choice of which objects to move to a buffer and which buffer to use. Alg. \ref{alg:NonMonotone} describes a non-monotone search for this purpose, which searches through the space of perturbations $P(o_i, b_i)$.

\begin{algorithm}
\DontPrintSemicolon
\begin{small}
    \SetKwInOut{Input}{Input}
    \SetKwInOut{Output}{Output}
    \SetKwComment{Comment}{\% }{}
    \caption{ Informed-Search($\arrangement_I$, $\arrangement_F$)}
		\label{alg:NonMonotone}
    \SetAlgoLined
		$T \leftarrow \emptyset$\\
        $T_{new}$ = $\tt \DFSDP$-$\tt LocalPlanner$($\arrangement_I$, $\arrangement_F$)\\
        \lIf{$\arrangement_F \in T_{new}.V$}{\Return $T_{new}$}
        $T\leftarrow T + T_{new}$\\
        \While{$T$ is not connected to $\arrangement_F$}{
        $\arrangement_C \leftarrow$ \textsc{Select-Expansion-Node}($T$)\\
        \lIf{$\arrangement_C = \emptyset$}{$\arrangement_C \leftarrow \textsc{RandomNode}(T.V)$}
        $\mathcal{O_C} \leftarrow$ \textsc{Select-Perturbation-Object}($\arrangement_C$)\\
        $\mathcal{B_C} \leftarrow$ \textsc{Select-Perturbation-Buffer}($\arrangement_C, \mathcal{O_C}$)\\
        \For{$P(o_i, b_i) \in (\mathcal{O_C}, \mathcal{B_C})$}{
        $T_{new}$ = $\tt \EDFSDP$-$\tt LocalPlanner(\arrangement_C, \arrangement_F, P(o_i,b_i)$)\\
        \lIf{$\arrangement_F \in$ $T_{new}.V$}{\Return $T+T_{new}$}
        \lElse{$T\leftarrow T + T_{new}$}
        }
        }
\end{small}
\end{algorithm}

The framework receives as input the problem instance ($\arrangement_I, \arrangement_F$) and returns a search tree in the arrangement space with a path from $\arrangement_I$ to $\arrangement_F$. It initializes the search tree $T$ (line 1) and calls the proposed monotone solver, which recursively calls Alg. \ref{alg:DFSDP}, to return a subtree $T_{new}$ rooted at $\arrangement_I$ (line 2) and identify if the problem is monotone (line 3). If the problem is not monotone, the partial solution (subtree) $T_{new}$ is added to the search tree (line 4) and new perturbations are attempted until a solution is found (line 5). For each perturbation, the algorithm decides on a new expansion node (the arrangement $\arrangement_C$ to launch the non-monotone local planner) (line 6) and which objects $\mathcal{O_C}$ (line 8) to place in which buffers $\mathcal{B_C}$ (line 9). For each computed perturbation $P(o_i, b_i)$ (line 10), the non-monotone planner (e.g. $\tt \EDFSDP$) attempts to solve the subproblem $\arrangement_C \rightarrow \arrangement_F$ with $P(o_i, b_i)$ (line 11) and a solution is returned if solved (line 12). Otherwise, a partial solution $T_{new}$ is added to the search tree (line 13). If no expansion node is recommended, a random node in the tree will be selected for potential perturbations (line 7) in order to ensure exhaustiveness. For the accompanying experiments, the search stops when a time threshold is exceeded.

As indicated in Alg.~\ref{alg:NonMonotone}, the key component of selecting a  promising perturbation involves, selecting an arrangement node $\arrangement_C$ to expand, deciding which objects to move to buffers, and selecting buffers to use.

\begin{figure}[ht]
    \centering
    \includegraphics[width=0.45\textwidth]{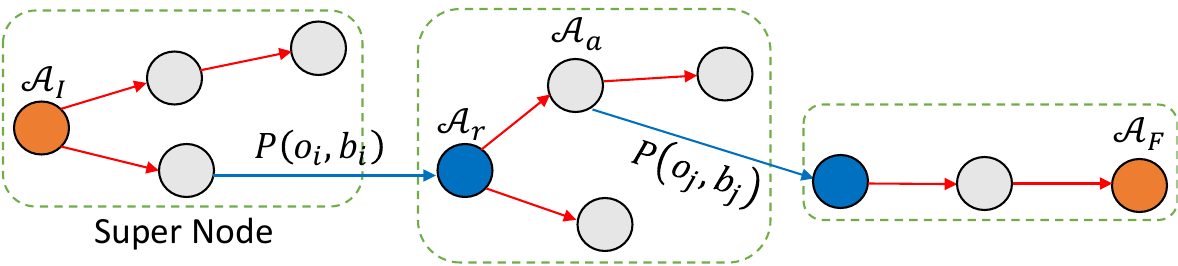}
    \caption{A search tree built according to Alg.~\ref{alg:NonMonotone}. Nodes correspond to arrangements, while directed edges correspond to monotone transitions from one arrangement to another. A super node is a set of nodes which are
    accessible
    from the same root; they are connected via a "perturbation" (blue arrow), where an object is moved to a buffer. Except for the initial arrangement $\arrangement_I$, roots of super nodes (blue circles) are perturbation nodes. }
    \label{fig:arrangement_search_tree}
\end{figure}

\subsection{Selecting Arrangements for Expansion}
\label{subsec:arrangement selection}


As shown in Fig.~\ref{fig:arrangement_search_tree}, the search tree can be divided into multiple subtrees, referred to as \emph{super nodes}, based on perturbations. Each super-node is rooted at $\arrangement_I$ or a \emph{perturbation node} (blue circles), which is defined as the arrangement derived from a perturbation. Given a super node rooted at $\arrangement$, other members of the super node are the nodes on the search tree which are reachable from $\arrangement$ via monotone transitions, i.e., without perturbations. In Fig. \ref{fig:arrangement_search_tree}, assuming a path from $\arrangement_a$ to $\arrangement_F$ can be found via the local planner $\tt \EDFSDP$ given a perturbation $P(o_j, b_j)$, then that path can also be found by calling $\tt \EDFSDP$ from $\arrangement_r$, the root of the super node that $\arrangement_a$ belongs in. If the solution path exists along some of the siblings of $\arrangement_a$, but is not discoverable from $\arrangement_a$, calling $\tt \EDFSDP$ at $\arrangement_r$ will still work. Therefore, for each super node, launching the planner from the root is the preferred way to generate solutions. Among the super-nodes, those with fewer perturbations relative to $\arrangement_I$ are prioritized. Among those with the same number of perturbations, those with more objects at their goal positions are prioritized as they are more similar to $\arrangement_F$.


\subsection{Selecting Objects for Perturbation}
\label{subsec:object_selection}

Once an arrangement is selected to be expanded, a decision needs to be made about which objects to move to a buffer. Intuitively, a highly \emph{constraining} object, whose current position $\arrangement_C[o]$ blocks other objects to make progress to their goals, should be prioritized. Symmetrically, a highly \emph{constrained} object, whose goal position $\arrangement_F[o]$ is blocked by other objects not at their goals, should not be considered to move directly towards its goal. In summary, the priority for selecting perturbations should be given to objects that are both highly constraining and constrained.

The extent to which an object is constraining or constrained can be measured via its degree in an approximate dependency graph, which uses the interference set $\mathbf{I}_{i,\arrangement}(p)$ of the current positions of objects not yet moved to their goals in $\arrangement_C$. In particular, denote $\overline{\mathcal{O}}(\arrangement_C)$ as the sets of objects not yet moved to their goals at $\arrangement_C$. If the current position $\arrangement_C[o_i]$ of an object $o_i \in \overline{\mathcal{O}}(\arrangement_C)$ interferes with the goal position $\arrangement_F[o_j]$ of another object $o_j \in \overline{\mathcal{O}}(\arrangement_C)$, then it creates a dependency edge $e(j,i)$, indicating that $o_j$ depends on object $o_i$. An indicator variable $\mathds{1}(j,i)$ is defined as 
\begin{equation*}
    \label{depedency_edge_indicator}
        \mathds{1}(j,i) =
        \begin{cases}
            1, & \text{if } V(\arrangement_C[o_i]) \cap V(\arrangement_F[o_j]) \neq \emptyset \\
            0, & \text{otherwise}.
        \end{cases} 
\end{equation*}
where $\mathds{1}(j,i) = 1$ indicates the existence of edge $e(j,i)$.

Given all such dependency edges, define $\mathcal{D}_I(o_i) = \sum_{o_j \in \overline{\mathcal{O}}(\arrangement_C)}\mathds{1}(j,i)$ as the \emph{inner degree of dependency} which indicates the degree of which the object $o_i$ is constraining at $\arrangement_C$ and $\mathcal{D}_O(o_i) = \sum_{o_j \in \overline{\mathcal{O}}(\arrangement_C)}\mathds{1}(i,j)$ as the \emph{outer degree of dependency} which indicates the degree of which the object $o_i$ is constrained at $\arrangement_C$. Then, all objects in $\overline{\mathcal{O}}(\arrangement_C)$ are  ranked in descending order of $\mathcal{D}_I(o_i) + \mathcal{D}_O(o_i)$,
so that the most constraining and constrained objects are prioritized.

\subsection{Selecting Buffers for Perturbation}
\label{subsec:buffer_selection}


Given an instance ($\arrangement_I, \arrangement_F$), candidate buffers are generated before the search. The process samples buffers overlapping with the fewest start and goal positions of objects, as well as of previously generated buffers. Besides the candidate buffers generated, unoccupied start and goal positions can also be used as buffers for some arrangement $\arrangement_C$. Given this observation, each object $o_i$ is assigned a buffer online by ranking candidates according to the following priorities. Buffers not overlapping with starts and goals are considered first as long as they are reachable from $\arrangement_C[o_i]$. Then, start positions of objects already at goals in $\arrangement_C$ are considered. Finally, goal positions not occupied in $\arrangement_C$ are considered.

\section{Experiments}
\label{sec:experiments}

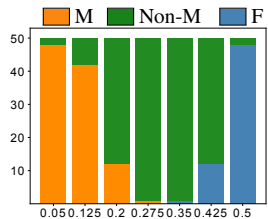
\begin{wrapfigure}[11]{l}{0.2\textwidth}
\vspace{1mm}
\begin{overpic}[width=0.2\textwidth]{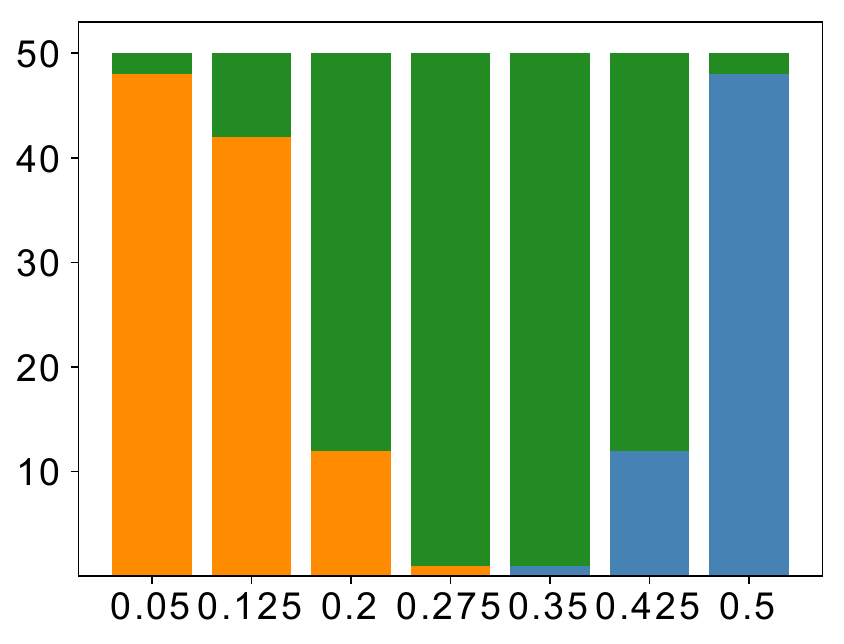}
\definecolor{Orange}{RGB}{252, 134, 15}
\put(14,78){\fcolorbox{black}{Orange}{ }}
\put(26, 75.5){{\footnotesize M}}
\definecolor{Green}{RGB}{26, 130, 26}
\put(38,78){\fcolorbox{black}{Green}{ }}
\put(50, 75.5){{\footnotesize Non-M}}
\definecolor{Blue}{RGB}{70, 130, 180}
\put(80,78){\fcolorbox{black}{Blue}{ }}
\put(92, 75.5){{\footnotesize F}}
	\end{overpic}
\vspace*{-6.5mm}
\caption{A distribution of monotone (M), non-monotone (Non-M), and failed (F) cases among 50 instances (y-axis) for increasing density (x-axis).}
\label{fig:density}
\end{wrapfigure}
To evaluate performance of the proposed algorithms, experiments on monotone and non-monotone instances are performed for different environment ``density'' levels. The density of the environment is defined as the ratio of the area occupied by objects to that of the environment.
For a fixed number of objects, the density level can be increased by either increasing the object sizes or decreasing the workspace area.
Fig. \ref{fig:density} 
shows the relationship between the probability of generating (or failing to generate) a monotone and non-monotone instance with 10 objects for different density levels. An attempt will fail when the environment has no more space for a valid start/goal placement. According to Fig. \ref{fig:density}, the density level of the monotone and non-monotone problem should be chosen in the ranges $[0, 0.2]$ and $[0.125, 0.425]$ respectively.

\subsection{Evaluation on Monotone Problems}
\label{subsec: experiment_monotone}
For monotone problems, $\tt \DFSDP$ is compared against several leading monotone solvers in the same field.
\begin{enumerate}[leftmargin=*]
    \item $\tt mRS$ (monotone rearrangement solver) - A backtracking method which searches over all possible orders with which the objects can be moved \cite{stilman2007manipulation}.
    \item $\tt fmRS$ (fast-$\tt mRS$) - Compute the sequence of moving objects via topological sorting on a constraint graph \cite{krontiris2016efficiently} constructed by assigning a transfer path per object.
    \item $\tt IP$-$\tt Solver$ - An integer programming (IP) solver which selects a path per object among all path options such that the corresponding dependency graph is acyclic.
\end{enumerate}

\begin{figure}[ht]
    \begin{center}
\begin{overpic}[scale=.27]{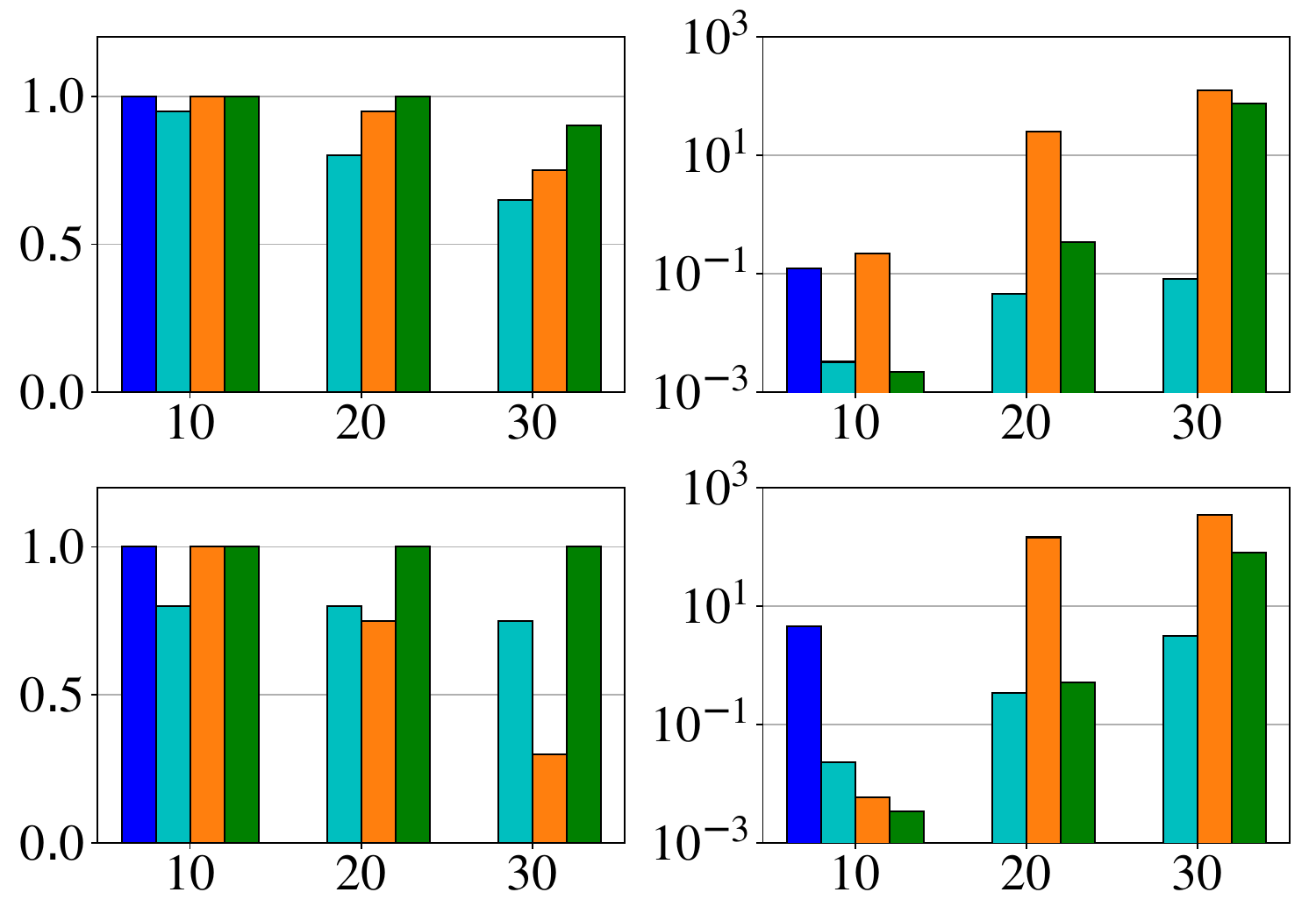}
\put(7,-1){\fcolorbox{black}{blue}{  }}
\put(13, -2.15){{\footnotesize $\tt IP$-$\tt Solver$}}
\definecolor{Green}{RGB}{0, 128, 0}
\put(34,-1){\fcolorbox{black}{cyan}{  }}
\put(40, -2.15){{\footnotesize $\tt fmRS$}}
\put(54,-1){\fcolorbox{black}{orange}{  }}
\put(60, -2.15){{\footnotesize $\tt mRS$}}
\put(74,-1){\fcolorbox{black}{Green}{  }}
\put(80, -2.15){{\footnotesize $\tt \DFSDP$}}
\end{overpic}
\end{center}
\vspace{-1mm}
\caption{\label{fig:monotone-experiments}[Top] Success rate (left) and computation time in seconds (right) with density level 0.1. [Bottom] The same evaluation with density level 0.2.
Number of objects are shown on the x-axis.}
\vspace{-1mm}
\end{figure}

\begin{figure*}[t]
    \centering
    \begin{tabular}{ccc}
        \includegraphics[width = 0.33 \textwidth]{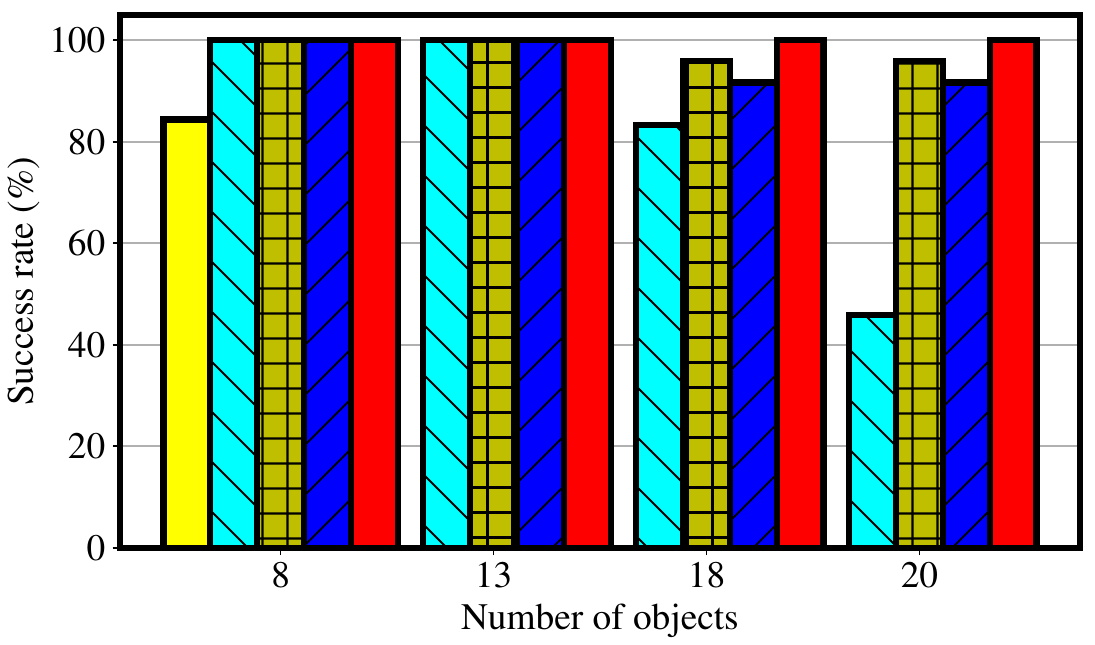} &
        \includegraphics[width = 0.33 \textwidth]{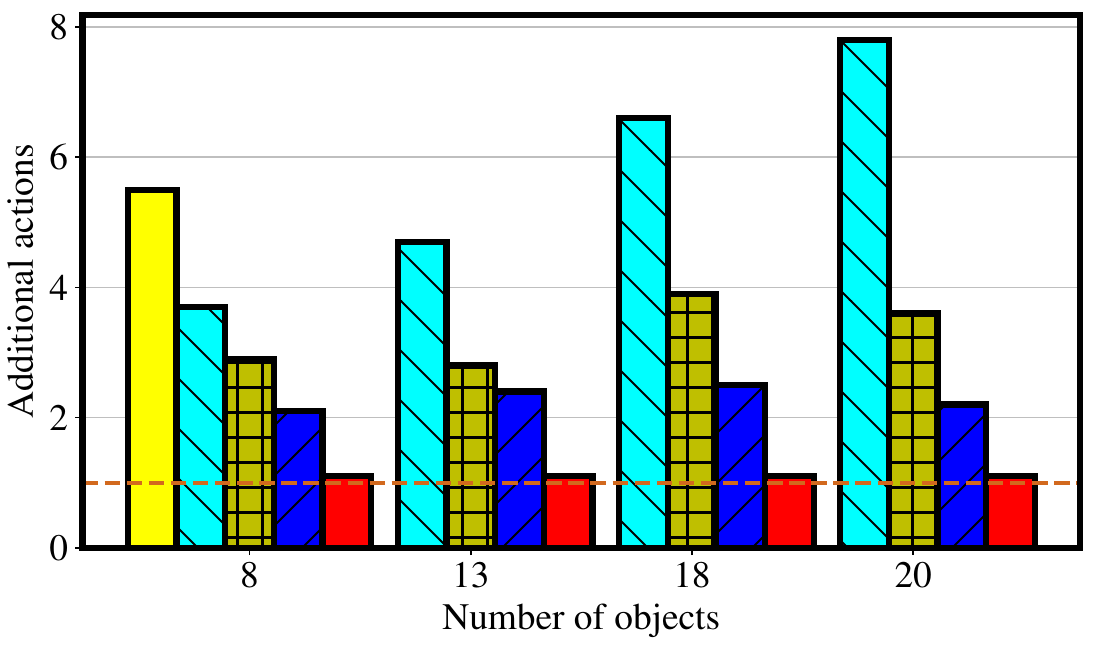} & 
        \includegraphics[width = 0.33 \textwidth]{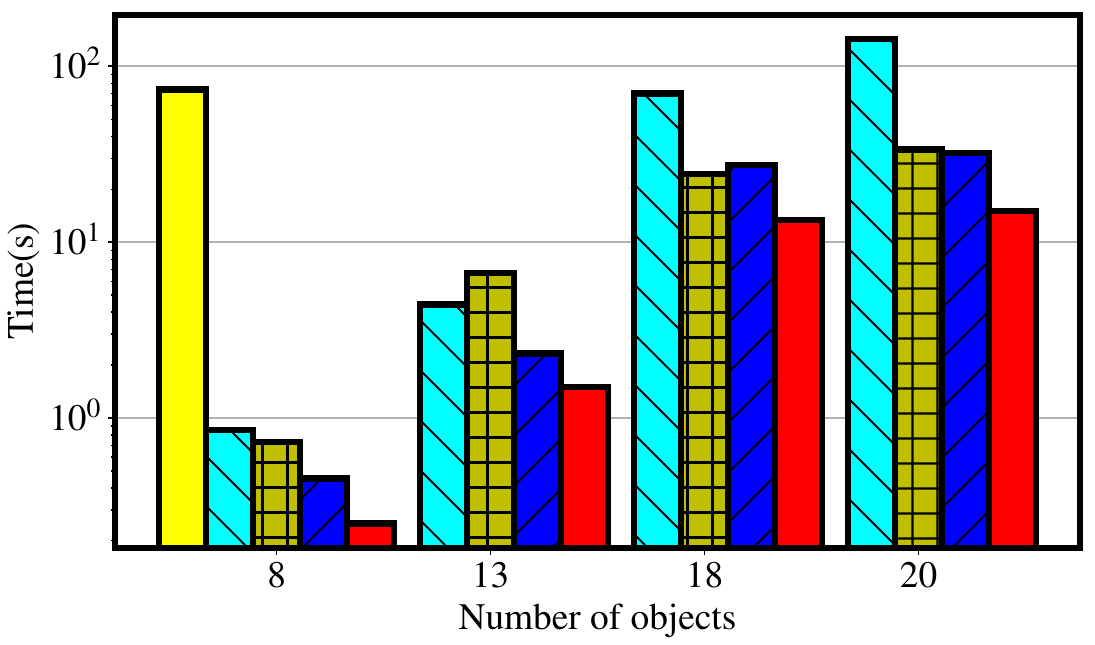} \\
                \includegraphics[width = 0.33 \textwidth]{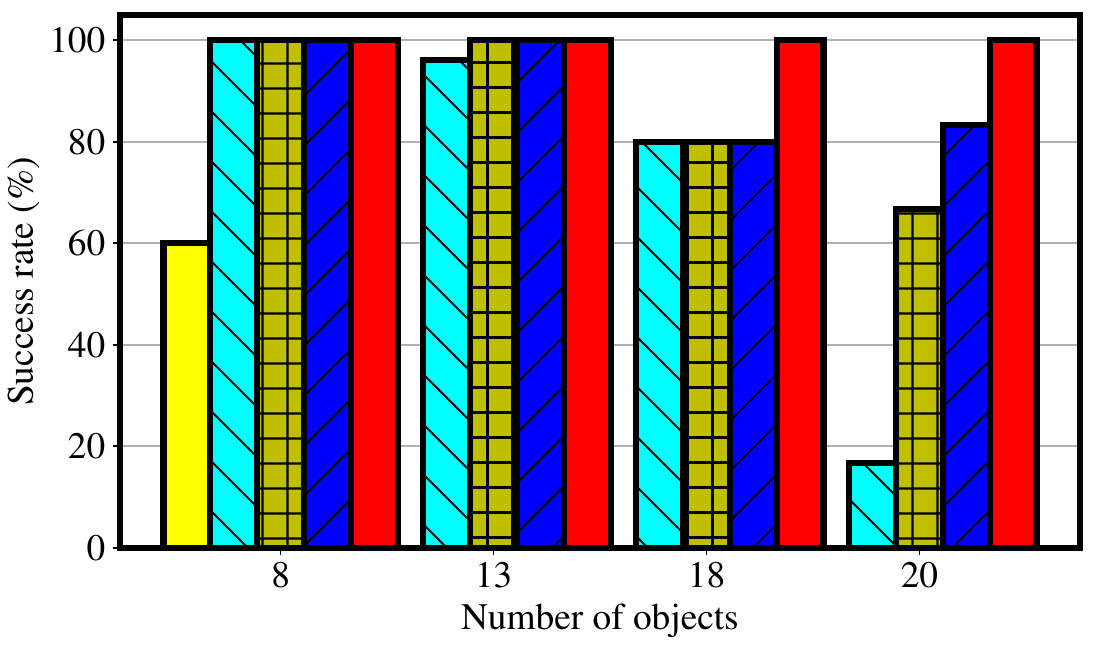} &
        \includegraphics[width = 0.33 \textwidth]{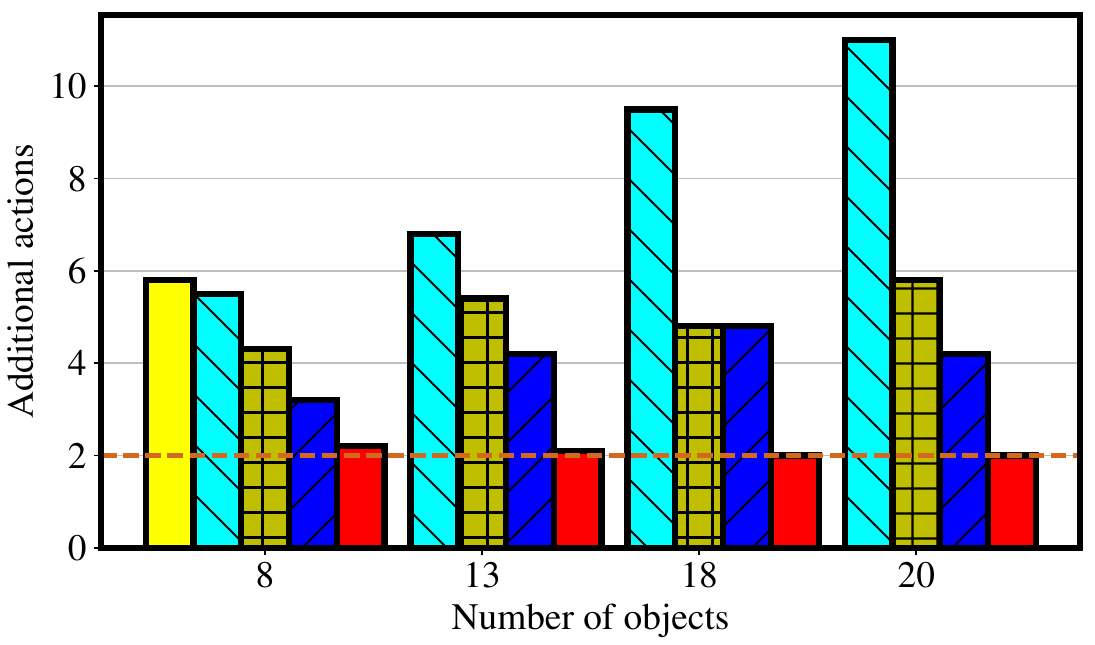} & 
        \includegraphics[width = 0.33 \textwidth]{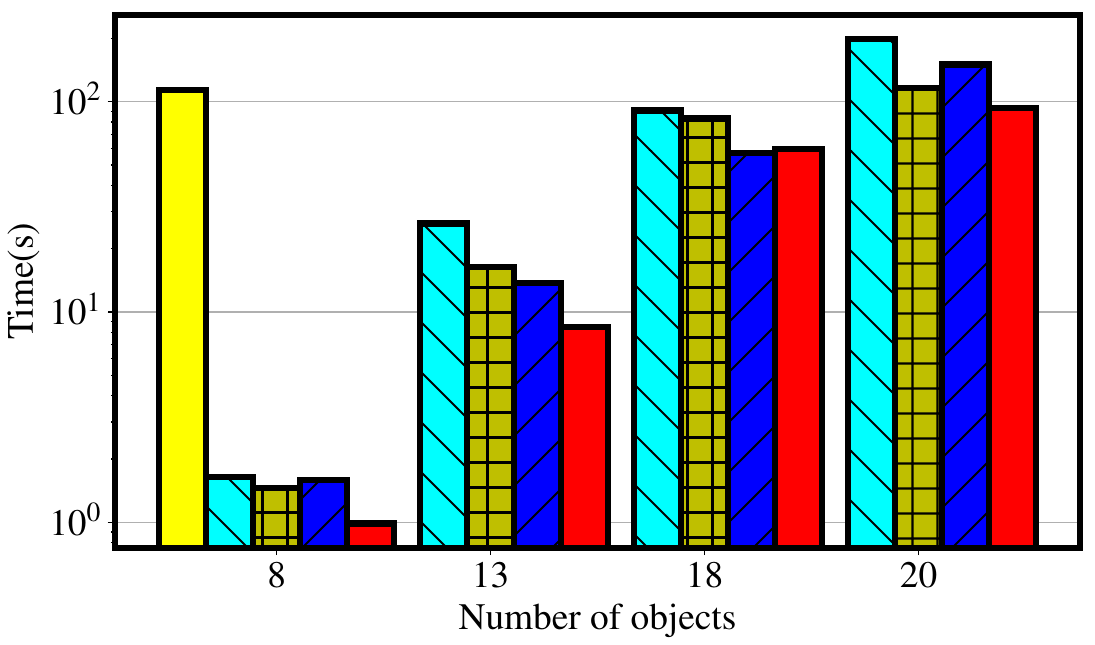} \\
    \end{tabular}
    \includegraphics[width = 0.95 \textwidth]{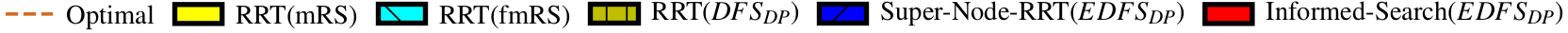}
    \vspace{-2mm}
    \caption{Experimental results on one-buffer (top row) and two-buffer (bottom row) instances evaluating (1) success rate on finding a solution (left column) (2) additional actions needed to solve the problem (middle column) and (3) computation time (right column).}
    \label{fig:non-monotone results with static buffer}
\end{figure*}


The experiments are conducted in a square environment with density level 0.1 and 0.2 and the number of objects ranges between 10 and 30. For each setup, 20 instances are generated. A planning attempt is considered a failure if the method cannot determine monotonicity within 500 seconds or return a negative response. The success rate (left) and average running time (right) is plotted in Fig. \ref{fig:monotone-experiments}.

The $\tt IP$-$\tt solver$ has to pre-compute all the region graph paths for each pair of start and goal object positions, from which to make a choice, thus making it unscalable beyond 10 objects. $\tt mRS$ considers all possible moving orders (time complexity of $O(n!)$) while the proposed $\tt \DFSDP$ keeps solving subproblems without considering the ordering of the solved subproblems ($O(2^n)$). Therefore, $\tt \DFSDP$ significantly outperforms $\tt mRS$ both in higher success rate and lower computation time as the number of objects increases. The $\tt fmRS$ greedily assigns one path to each object with the minimum constraints violated, which indicates incompleteness compared to the complete $\tt \DFSDP$. As a result, it is fast but tends not to find valid solutions in monotone instances.

\subsection{Evaluation on Non-monotone Problems}
\label{subsec: experiment_nonmonotone}
For non-monotone problems, the focus is on cases where the optimal solution requires one or two buffers and can be computed with an expensive, exhaustive brute force search approach. The density level is set to 0.225. The following comparison points are considered:
\begin{enumerate}[leftmargin=*]
    \item $\tt RRT(mRS)$ - Searches the arrangement space in an ($\tt RRT$)-like fashion \cite{lavalle1998rapidly} and uses $\tt mRS$ as a local planner.
    \item $\tt RRT(fmRS)$ - Same as above but $\tt fmRS$ as the local planner.
    \item $\tt RRT(\DFSDP)$ - Same but $\tt \DFSDP$ as the local planner.
    \item $\tt Super$-$\tt Node$-$\tt RRT(\EDFSDP)$ - Select the roots of super nodes for expansion and uses $\tt \EDFSDP$ as the local planner.
    \item $\tt Informed$-$\tt Search(\EDFSDP)$ - Same as above but uses heuristics to order super nodes, objects, and buffers. 
\end{enumerate}

Fig.~\ref{fig:non-monotone results with static buffer} demonstrates the results for 1 and 2 buffer cases respectively in terms of success rate in finding a feasible solution in 300 seconds, number of buffers used and computation time. 
The two baselines $\tt RRT(\DFSDP)$ and $\tt Super$-$\tt Node$-$\tt RRT(\EDFSDP)$ search the arrangement space using an $\tt RRT$-like process for sampling and therefore focus on finding feasible solutions, instead of high-quality ones. The total number of buffers is much higher than the final informed search pipeline (middle column) as the proposed informed search prioritizes super nodes with fewer number of perturbation to expand. In addition, the success rate starts to drop (left column) and the computation time starts to increase (right column) for the baselines as the number of objects increases, since the arrangement space increases exponentially and random perturbation node selection performance suffers. The improved baseline $\tt Super$-$\tt Node$-$\tt RRT(\EDFSDP)$ outperforms $\tt RRT(\DFSDP)$ by always selecting the root of super nodes and utilizing the benefits of $\tt \EDFSDP$ as a local planner. The proposed informed framework focuses the search to the most promising part of arrangement space and explores it more systematically. Therefore, it finds near-optimal solutions (middle column: 1.07 in one-buffer cases, 2.09 in two-buffer cases) with $100\%$ success rate even in harder instances (left column: 18, 20 objects with 2 buffers). The computation time is lower than other methods in one-buffer cases and remains competitive in two-buffer cases. 


The comparison results with $\tt RRT(mRS)$ and $\tt RRT(fmRS)$ are consistent with the observation made in monotone evaluation. The solution quality of $\tt RRT(fmRS)$ is worse than the baseline version $\tt RRT(\DFSDP)$ of the proposed search framework due to the incompleteness of the local solver. In addition, $\tt RRT(mRS)$ is not as scalable in non-monotone instances due to the weakness in quickly identifying non-monotonicity. 


\section{Conclusion and Future Work}
This work tackles uniform object rearrangement with the goal of minimizing the total number of object movements. A region graph is introduced to decompose the configuration space and classify all continuous paths, upon which a complete and efficient monotone solver and then an effective informed search framework for non-monotone problems are proposed. The framework achieves high-quality solutions with high success rate and reduced computation time relative to alternatives.

The current study motivates future work for these efficient tools, including the extension to object-robot interactions and generalizations to non-uniform object geometries. Furthermore, the region graph can also be exploited to dynamically generate buffers, which adapts to specific arrangement environments to improve performance. It would be interesting to show (or disprove) that it is sufficient to consider a finite set of buffers for guaranteeing a complete approach. Machine learning methods can also be used to learn heuristics and object dependencies given access to solutions by this planning approach. This can lead to even more scalable solutions. Perception should also be taken into account to reason about pose hypotheses per object, which could affect the arrangement ordering, such as moving an object to make another more discernible. Different types of manipulation primitives can also be considered, such as non-prehensile actions (e.g., pushing, flipping) to enrich rearrangement strategies, which can be generalized to more sophisticated environments.


\bibliographystyle{format/IEEEtran}
\bibliography{bib/c}

\end{document}